# GLCM based Chi-square Histogram Distance for Automatic Detection of Defects on Patterned Textures

V. Asha
Department of Computer Applications
New Horizon College of Engineering
Bangalore, Karnataka, INDIA
(Research Scholar, JSS Research Foundation, SJCE Campus, University of Mysore, Mysore, Karnataka, India)
v_asha@live.com

N. U. Bhajantri
Department of Computer Science Engineering
Government Engineering College
Chamarajanagar, Mysore District, Karnataka, INDIA
bhajan3nu@gmail.com

P. Nagabhushan
Department of Studies in Computer Science
Mysore University, Mysore, Karnataka, INDIA
pnagabhushan@hotmail.com

***Abstract*— Chi-square histogram distance is one of the distance measures that can be used to find dissimilarity between two histograms. Motivated by the fact that texture discrimination by human vision system is based on second-order statistics, we make use of histogram of gray-level co-occurrence matrix (GLCM) that is based on second-order statistics and propose a new machine vision algorithm for automatic defect detection on patterned textures. Input defective images are split into several periodic blocks and GLCMs are computed after quantizing the gray levels from 0-255 to 0-63 to keep the size of GLCM compact and to reduce computation time. Dissimilarity matrix derived from chi-square distances of the GLCMs is subjected to hierarchical clustering to automatically identify defective and defect-free blocks. Effectiveness of the proposed method is demonstrated through experiments on defective real-fabric images of 2 major wallpaper groups (pmm and p4m groups).**

## I. Introduction

The need for automated defect detection scheme in industries stems from quality control of industrial products. Quality control is a key aspect in various industries. Conventional human-vision based inspections demand for high labor cost and skilled inspectors. Moreover, in the conventional human-vision based inspections, lack of repeatability and reproducibility of inspection results due to fatigue and subjective nature of human inspections and imperfect defect detection are common. These, in turn, affect the inspection quality and the production rate. An automated inspection system can help in reducing the inspection time and increasing the production rate. Among various industries, textile industry is one of the biggest traditional industries requiring automated inspection. Patterned textures are often found in various applications such as ceramic tiles, wallpapers, and textile fabrics. Modern textile industries produce so many varieties in fabric design. Nevertheless, all patterned fabrics produced by the modern textile industries can be grouped into only 17 wallpaper groups which are denoted as p1, p2, p3, p3m1, p31m, p4, p4m, p4g, pm, pg, pmg, pgg, p6, p6m, cm, cmm and pmm [1], [2]. These wallpaper groups basically consist of lattices of parallelogram, rectangular, rhombic, square or hexagonal shape. A wallpaper group has at least one of the characteristics of translational, rotational, reflectional and glide-reflectional symmetries. In fact, p1 defines a texture with just one lattice repeating itself over the complete image such as plain and twill fabrics as shown in Fig. 1 (a) and (b). Among the other 16 Wallpaper groups, pmm, p2 and p4m are called major wallpaper groups due to the fact that the wallpaper groups other than pmm, p2 and p4m can be transformed into these 3 major groups through geometric transformation [3]. Examples of fabrics belonging to these major wallpaper groups are shown in Fig. 1 (c), (d) and (e). Due to complexity in the design, existence of numerous categories of patterns, and similarity between the defect and background, most of the methods in literature depend on training stage with numerous defect-free samples for obtaining decision-boundaries or thresholds prior to defect detection [4] – [8]. In this paper, we propose a method of defect detection on patterned fabric images without any training stage with the help of texture-periodicity and chi-square histogram distance derived from gray level co-occurrence matrices. The main contributions of this research can be summarized as follows:

- The proposed method of defect detection is more generic as the method can be applied to periodic images belonging to 16 out of 17 wallpaper groups (other than p1 group images such as plain and twill fabric images).
- The proposed method does not require any training stage with defect-free samples for decision boundaries or

thresholds unlike other methods. As a result, the proposed method does not need huge memory space for storage of defect-free samples.
- Detection of defective and defect-free periodic units is automatically carried out based on cluster analysis without human intervention.

The program for the proposed algorithm is written in Matlab-7.0 and run in a Pentium-IV Personal Computer of RAM capacity 2 GB. The organization of this paper is as follows: Section-II presents a brief review on chi-square histogram based on gray level co-occurrence matrices and the proposed algorithm for defect detection along with illustration and results from experiments on various real fabric images with defects. Section-III has the conclusions.

## II. Proposed Method of Defect Detection

Chi-square distance is one of the distance measures that can be used as a measure of dissimilarity between two histograms and has been widely used in various applications such as image retrieval, texture and object classification, and shape classification [9]. In histograms of many processes, the difference between large bins is less important than the difference between small bins and that should be reduced. The chi-square histograms take this into account [9]. The chi-square histogram distance comes from the chi-square statistics to test the fit between a distribution and observed frequencies. Based on the findings from [10] that human vision perception for texture discrimination is based on second-order statistics, we make use of histogram of gray-level co-occurrence matrix (GLCM) that is based on second-order statistics to discern between defective and defect-free periodic block. Since the time of proposal of 14 features derived from GLCM by Haralick *et al.* [11], several authors have used GLCM features for various texture analysis applications (eg. [12]–[16]). In this paper, we intend to use information from every pixel-pair of a periodic block with that of other periodic block to compute the distance metric. The uniqueness of this distance metric based on GLCM is that the distance measure is considered for every pair of gray levels between two periodic blocks unlike the conventional GLCM-features that are computed for the entire GLCM.

### A. Chi-square Histogram based on GLCM

Gray level co-occurrence matrices have been employed in extracting texture features for various applications for a long time. It estimates the gray-level dependencies in a local neighborhood for a given pixel displacement and orientation. In other words, a GLCM is a matrix that counts the number of times a pixel with grey-level $i$ occurs at position a vector from a pixel with grey-level $j$. Mathematically, the GLCM for an image $I(x, y)$ of size $(M, N)$ parameterized by the offset $(\Delta x, \Delta y)$ is given as [11]

$$C_{\Delta x,\Delta y}(i,j)=\sum_{x=1}^{M}\sum_{y=1}^{N}\begin{cases}1, \text{if } I(x,y)=i \\ \text{and } I(x+\Delta x, y+\Delta y)=j \\ 0, \text{otherwise}\end{cases} \tag{1}$$

The offset $(\Delta x, \Delta y)$ characterizes the pixel displacement and the orientation for which the co-occurrence matrix is calculated. If $p$ and $q$ represent the probability distributions of two events $A$ and $B$ with random variables, $i = 1, 2,\ldots, n$, the chi-square measure between these two histograms is given by [9]

$$\chi^2_{A,B}=\frac{1}{2}\sum_{i=1}^{n}\frac{[p(i)-q(i)]^2}{p(i)+q(i)} \tag{2}$$

The square root of chi-square measure is a distance metric [9] which is termed here as *Chi-square Histogram Distance*. If $p(i, j)$ and $p(i, j)$ represent the joint-probability histograms of two different images $A(M, N)$ and $B(M, N)$ respectively, with the dynamic range of gray values being $(0, L–1)$ (where $L$ is the total number of gray levels), the chi-square distance metric between these GLCM two histograms can be calculated from the following relation:

$$\chi_{A,B}=\sqrt{\frac{1}{2}\sum_{i=0}^{L-1}\sum_{j=0}^{L-1}\frac{[p(i,j)-q(i,j)]^2}{p(i,j)+q(i,j)}} \tag{3}$$

### B. Algorithm Description

There are three main assumptions in the proposed algorithm as follows:

- Test image is of at least two periodic units in horizontal direction and two in vertical direction whose dimensions are known a priori.
- Number of defective periodic units is always less than the number of defect-free periodic units.
- Test images are from imaging system oriented perpendicular to the surface of the product such as textile fabric. This assumption is due to the fact that in a defect detection system in industries such as fabric industry, the imaging system is always oriented perpendicular to the plane of fabric surface.

Based on our earlier approach of analyzing patterned textures [17], four cropped images are obtained from the defective test image by cropping it from all four corners (top-left, bottom-left, top-right and bottom-right). If $g$ is the an image of size $M \times N$ with row periodicity $P_r$ (i.e., number of columns in a periodic unit) and column periodicity $P_c$ (i.e., number of rows in a periodic unit), then the size of cropped image $g_{crop}$ is $M_{crop} \times N_{crop}$ where $M_{crop}$ and $N_{crop}$ are measured from top-left, bottom-left, top-right and bottom-right corners and are given by the following equations:

$$M_{crop}=\text{floor}(M/P_c)\times P_c \tag{4}$$

$$N_{crop}=\text{floor}(N/P_r)\times P_r \tag{5}$$

Each cropped image is split into several periodic blocks of size $P_c \times P_r$. In order to construct GLCMs, the number of gray levels is an important factor in the computation of GLCMs. The more levels included in the computation, the more accurate the extracted textural information, with, of course, a subsequent increase in computation costs. Too less gray levels will result in loss of information due to quantization. In general, the effect of *false-contouring* starts predominating in an image if the gray levels are quantized below the dynamic range 0–63 [18]. In the proposed method, each 8-bit test image having a dynamic range 0–255 is linearly quantized to 6-bit image having a dynamic range 0–63. Sum of GLCMs over 8 directions $\theta \in \{0, \pi/4, \pi/2, 3\pi/4, \pi, 5\pi/4, 3\pi/2, \text{ and } 7\pi/4\}$ for a unit pixel displacement that is rotation-invariant is utilized to get a distance matrix $\psi$ (i.e., dissimilarity matrix) containing chi-square distance metrics for each periodic block with respect to itself and all other periodic blocks using (3) as below:

$$\Psi = \begin{bmatrix} \chi_{1,1} & \chi_{1,2} & \cdots & \chi_{1,n-1} & \chi_{1,n} \\ \chi_{2,1} & \chi_{2,2} & \cdots & \chi_{2,n-1} & \chi_{2,n} \\ \vdots & \vdots & \ddots & \vdots & \vdots \\ \chi_{n-1,1} & \chi_{n-1,2} & & \chi_{n-1,n-1} & \chi_{n-1,n} \\ \chi_{n,1} & \chi_{n,2} & \cdots & \chi_{n,n-1} & \chi_{n,n} \end{bmatrix} \quad (6)$$

It should be noted that if a cropped image has $n$ number of periodic blocks, then the size of the dissimilarity matrix is $n \times n$. Since dissimilarity of a periodic block with itself is zero and dissimilarity between $i^{th}$ periodic block and $j^{th}$ periodic block is same as that between $j^{th}$ periodic block and $i^{th}$ periodic block, the dissimilarity matrix becomes a diagonally symmetric matrix hollow matrix as below:

$$\Psi = \begin{bmatrix} 0 & & & & \\ \chi_{2,1} & 0 & & & \\ \vdots & \vdots & \ddots & & \\ \chi_{n-1,1} & \chi_{n-1,2} & & 0 & \\ \chi_{n,1} & \chi_{n,2} & \cdots & \chi_{n,n-1} & 0 \end{bmatrix} \quad (7)$$

It may be noted that because the matrix is similar about the diagonal, the upper diagonal elements are not filled for the sake of simplicity. This dissimilarity matrix is directly given as input to the Ward's cluster algorithm [19] to automatically get defective and defect-free periodic blocks from each cropped image. Detection of defective periodic blocks from each cropped image does not give an overview of the total defects in the input defective image. Hence, in order to get the overview of the total defects in the input image, we use *defect-fusion* proposed in [17] that involves merging of the boundaries of the defective periodic blocks identified from each cropped image, morphological filling [18] and Canny edge extraction [18].

### *C. Algorithm Illustration with Example*

In order to illustrate the proposed algorithm for defect detection, let us consider a pmm defective dot-patterned fabric image as shown in Fig. 2 (a). Following (4) and (5), four cropped images containing complete number of periodic blocks are obtained from the test image with the help of periodicities known a priori. Each cropped image is split into several periodic blocks and dissimilarity matrices are obtained based on (7). These dissimilarity matrices are shown in Fig. 2 (b)-(e) in gray-scale form by scaling the matrix elements linearly in the range 0–255. It may be noted from Fig. 2 (b)-(e) that the diagonal elements in the dissimilarity matrix indicate that the periodic blocks are of zero dissimilarity with themselves and that the dissimilarity matrix is symmetric. The dendrograms obtained from hierarchical clustering of the dissimilarity matrices are shown in Fig. 3 along with the identified defective blocks. The defective blocks thus identified from each cropped image are shown in Fig. 4, where the boundaries of the defective blocks are highlighted using white pixels. The boundaries of defective periodic blocks identified from each cropped image are shown in Fig. 5 (a) by superimposing on the original defective image and in Fig. 5 (b) separately on plain background. The morphologically filled zones are shown in Fig. 5 (c) and the edges extracted using Canny's edge operator are shown superimposed on original defective image in Fig. 5 (d). Thus, it is clear that *fusion* of defects from all 4 cropped images helps in getting an overview of total defects in the input image.

### *D. Experiements on 2 Major Wallpaper Groups*

As far as defect detection based on lattice concept is concerned, reason behind choosing pmm, p2, and p4m wallpaper groups is that all other wallpaper groups can be transformed into these 3 wallpaper groups through geometric transformations [3]. However, since the proposed method needs only horizontal and vertical periods, there is no need for geometric transformation of wallpaper groups other than pmm, p2 and p4m groups into pmm, p2 and p4m groups. Due to lack of required database, only pmm and p4m images with defects such as broken end (BE), thin bar (TNB), and thick bar (TKB) (as shown in Fig. 6) are utilized to show the effectiveness of the proposed algorithm. Defective periodic blocks identified from each cropped image of the defective fabric images and final results after merging of defects, morphological filling and edge detection are shown in Fig. 7.

### *E. Performance Evaluation of the Proposed Algorithm*

In order to access the performance of the proposed method, performance parameters, namely, precision, recall and accuracy [20], [21] are all evaluated in terms of true positive (TP), true negative (TN), false positive (FP), and false negative (FN), where true positive refers to the number of defective periodic blocks identified as defective, true negative is defined as the number of defect-free periodic blocks identified as defect-free, false positive refers to the number of defect-free periodic blocks identified as defective and false negative refers to the number of defective periodic blocks identified as defect-free. Precision is the number of periodic blocks correctly labeled as belonging to the positive class divided by the total number of periodic blocks labeled as belonging to the positive class and is calculated as TP/(TP+FP). Recall is the number of true positives divided by the sum of true positives and false negatives that are periodic blocks not labeled as belonging to the positive class but should have been and is calculated as TP/(TP+FN). Accuracy is the measure of success rate that considers detection rates of defective and defect-free periodic blocks and is calculated as (TP+TN)/(TP+TN+FP+FN). Though the number of periodic blocks from a defective input image is same for all of its cropped images, the number of defective periodic blocks identified does not have to be same for all cropped images. This is mainly due to the fact that the contribution of defect in each periodic block may differ for different cropped images. The performance parameters averaged over all cropped images for each defective image are given in Table 1 for pmm and p4m groups. The performance parameters averaged for all images of pmm and p4m wallpaper groups (viz., precision, recall and accuracy) are (100%, 82.2%, and 96.7%) and (100%, 81.6%, and 99.2%) based on total number of periodic samples – 756 and 1080 for pmm and p4m wallpaper groups respectively. It may be noted that relatively less recall rates indicate that there are few false negatives identified by the proposed method. However, because the proposed method yields high precision and accuracy, it can contribute to automatic defect detection in fabric industries.

TABLE I. SUMMARY OF PERFORMANCE PARAMETERS FOR EACH DEFECTIVE IMAGE (NOTE: BE=BROKEN END, TNB=THIN BAR, AND TKB=THICK BAR)

| Wallpaper group | Defect | No. of periodic blocks | Precision (%) | Recall (%) | Accuracy (%) |
|---|---|---|---|---|---|
| pmm | BE | 252 | 100 | 80.0 | 96.8 |
| | TNB | 252 | 100 | 75.0 | 98.4 |
| | TKB | 252 | 100 | 91.7 | 97.6 |
| p4m | BE | 360 | 100 | 91.7 | 99.4 |
| | TNB | 360 | 100 | 90.6 | 99.2 |
| | TKB | 360 | 100 | 62.5 | 98.9 |

## III. CONCLUSIONS

Through experiments on real fabric images of 2 major wallpaper groups (pmm and p4m) with defects, we have shown that the chi-square histogram based feature employing gray-level co-occurrence matrix is effective in indentifying fabric defects. Absence of training stage with defect-free samples for decision-boundaries or thresholds, unsupervised method of identifying defects using cluster analysis, and high success rates are the novelties of the proposed method. Thus, the proposed method can contribute to the development of computerized defect detection in fabric industries.

## ACKNOWLEDGMENT

The authors would like to thank Dr. Henry Y. T. Ngan, Research Associate of Industrial Automation Research Laboratory, Department of Electrical and Electronic Engineering, The University of Hong Kong, for providing the database of patterned fabrics.

## REFERENCES

[1] D. Schattschneider, “The Plane Symmetry Groups: Their Recognition and Notation,” Am. Math. Monthly, vol. 85, 1978, pp. 439–450.

[2] H. S. M. Coxeter and W. O. J. Moser, Generators and Relations for Discrete Groups, Fourth Edition, Springer–Verlag, New York, 1980.

[3] H. Y. T. Ngan, G. K. H. Pang and N. H. C. Yung, “Performance Evaluation for Motif-Based Patterned Texture Defect Detection,” IEEE Trans. on Autom. Sci. and Eng., vol. 7, no. 1, Jan. 2010, pp. 58–72.

[4] H. Y. T. Ngan and G. K. H. Pang, “Regularity Analysis for Patterned Texture Inspection,” IEEE Trans. on Autom. Sci. and Eng., vol. 6, no. 1, Jan. 2009, pp. 131–144.

[5] H. Y. T. Ngan and G. K. H. Pang, “Novel method for patterned fabric inspection using Bollinger bands,” Opt. Eng., vol. 45, no. 8, Aug. 2006, pp. 087202–1–15.

[6] F. Tajeripour, E. Kabir, and A. Sheikhi, “Fabric Defect Detection Using Modified Local Binary Patterns,” Proc. of the Int. Conf. on Comput. Intel. and Multiméd. Appl., vol. 2, Dec. 2007, pp. 261–267.

[7] H. Y. T. Ngan, G. K. H. Pang and N. H. C. Yung, “Motif-based defect detection for patterned fabric,” Pattern Recognit., vol. 41, 2008, pp. 1878–1894.

[8] H. Y. T. Ngan, G. K. H. Pang and N. H. C. Yung, “Ellipsoidal decision regions for motif-based patterned fabric defect detection,” Pattern Recognit., vol. 43, 2010, pp. 2132–2144.

[9] O. Pele and M. Werman, “The Quadratic-Chi Histogram Distance Family,” in K. Daniilidis, P. Maragos, N. Paragios, Eds.: Berlin, Heidelberg: ECCV 2010, Part II, LNCS 6312, 2010, pp. 749–762.

[10] T. Caelli, B. Julesz, and E. Gilbert, “On Perceptual Analyzers Underlying Visual Texture Discrimination: Part II,” Biol. Cybern., vol. 29, no. 4, 1978, pp. 201–214.

[11] R. M. Haralick, K. Shanmugam, and I. Dinstein, “Textural features for image classification,” IEEE Trans. on Syst., Man and Cybern., vol. 3, no. 6, 1973, pp. 610–662.

[12] A. L. Amet, A. Ertüzün, and A. Erçil, “Texture defect detection using subband domain co-occurrence matrices,” Image and Vis. Comp., vol. 18, 2000, pp. 543–553.

[13] J. C. A. Fernandos, J. A. B. C. Neves, and C. A. C. Couto, “Defect Detection and Localization in Textiles using Co-occurrence Matrices and Morphological Operators,” IEEE Int. Symp. on Ind. Electron., Montreal, Quebec, Canada, July 9–13, 2006.

[14] R. M. Haralick, “Statistical and structural approaches to texture,” Proc. of the IEEE, vol. 67, no. 5, 1979, pp. 786–804.

[15] C. F. J. Kuo and T. L. Su, “Gray Relational Analysis for Recognizing Fabric Defects,” Textile Res. J., vol. 73, no. 5, 2003, pp. 461–465.

[16] L. H. Soh, and C. Tsatsoulis, “Texture Analysis of SAR Sea Ice Imagery Using Gray Level Co-occurrence Matrices,” IEEE Trans. on Geosci. and Remote Sens., vol. 37, no. 2, Mar. 1999, pp. 780–795.

[17] V. Asha, N. U. Bhajantri, and P. Nagabhushan, “Automatic Detection of Texture Defects using Texture-Periodicity and Gabor Wavelets,” in K. R. Venugopal and L. M. Patnaik, Eds. Berlin, Heidelberg: ICIP 2011, CCIS 157, 2011, pp. 548–553.

[18] R. C. Gonzalez and R. E. Woods, Digital Image Processing, Third Edition, Pearson Prentice Hall, New Delhi, 2008.

[19] S. Theodoridis, K. Koutroumbas: Pattern Recognition, Fourth Edition, Academic Press, CA, 2009.

[20] T. Fawcett, “An introduction to ROC analysis,” Pattern Recognit. Lett., vol. 27, 2006, pp. 861–874.

[21] C. D. Brown and H. T. Davis, “Receiver operating characteristics curves and related decision measures: A tutorial,” Chemom. and Intell. Lab. Syst., vol. 80, 2006, pp. 24–38.

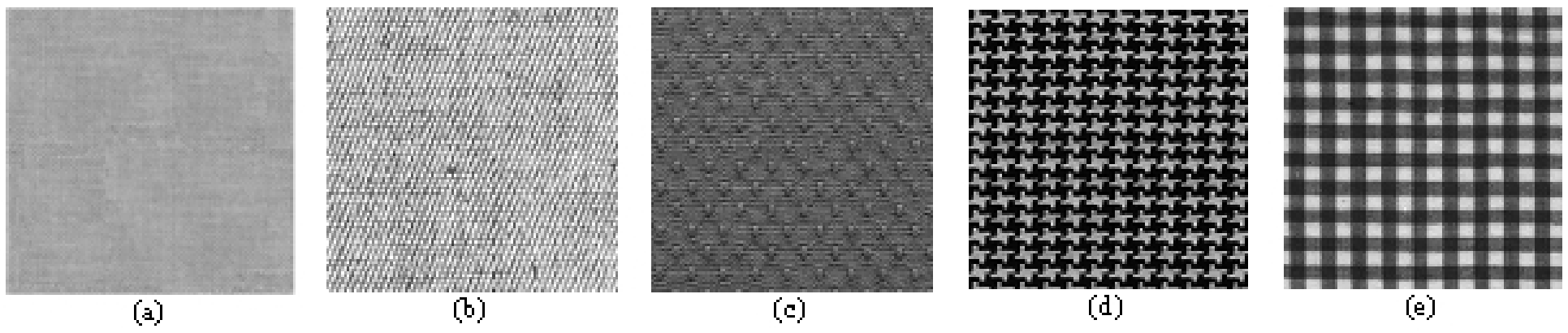

Figure 1. Fabric examples: (a) Plain fabric (p1 group); (b) Twill fabric (p1 group); (c) Dot-patterned fabric (pmm group); (d) Star-patterned fabric (p2 group); (e) Box-patterned fabric (p4m group).

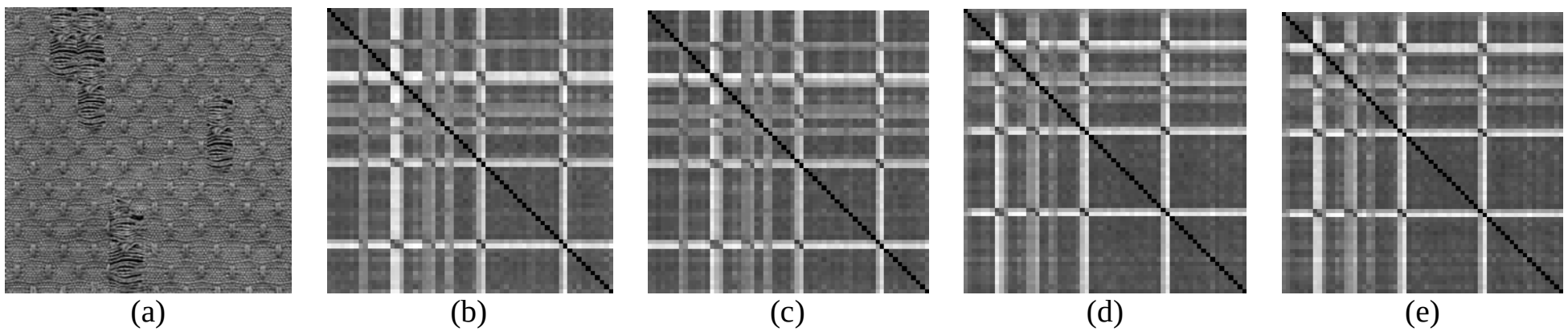

Figure 2. (a) Input defective image; (b), (c), (d) and (e) show the dissimilarity matrices derived from the chi-square distance metrics of the cropped image obtained from top-left, bottom-left, top-right and bottom-right corners of the test image respectively.

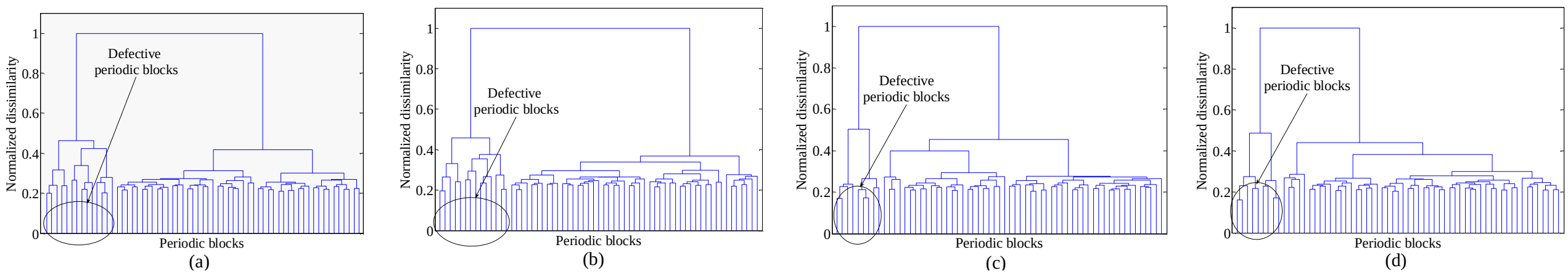

Figure 3. Dendrogram obtained from cluster analysis of chi-square dissimilarity matrix obtained from the test image by cropping it from (a) top-left (b) bottom-left (c) top-right and (d) bottom-right corners. Defective blocks identified from these cropped images are (8, 9, 15, 16, 17, 22, 23, 24, 27, 28, 34, 35, 52, and 53), (8, 9, 15, 16, 17, 22, 23, 24, 27, 28, 34, 35, 52, and 53), (8, 9, 15, 16, 17, 27, 28, 45, and 46) and (8, 9, 15, 16, 17, 27, 28, 45, and 46) respectively. It may be noted that since the cropped images have more number of periodic blocks, the periodic block identities in the abscissa are not shown in order to have better clarity.

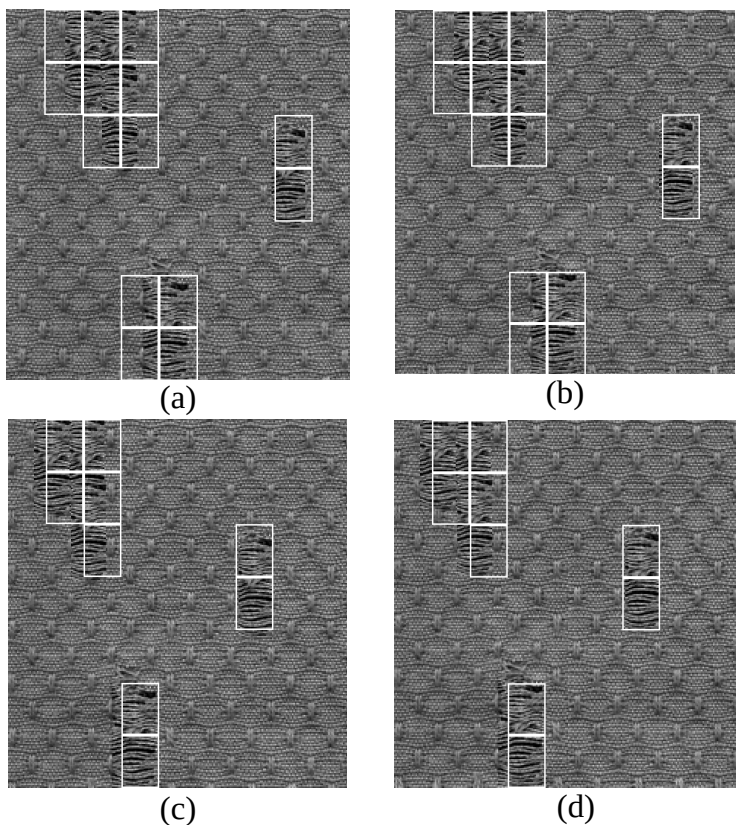

Figure 4. Defective periodic blocks identified from the cluster analysis of dissimilarity matrix derived from the chi-square distance metrics of the cropped images obtained from (a) top-left (b) bottom-left (c) top-right and (d) bottom-right corners of the test image with their boundaries highlighted using white pixels.

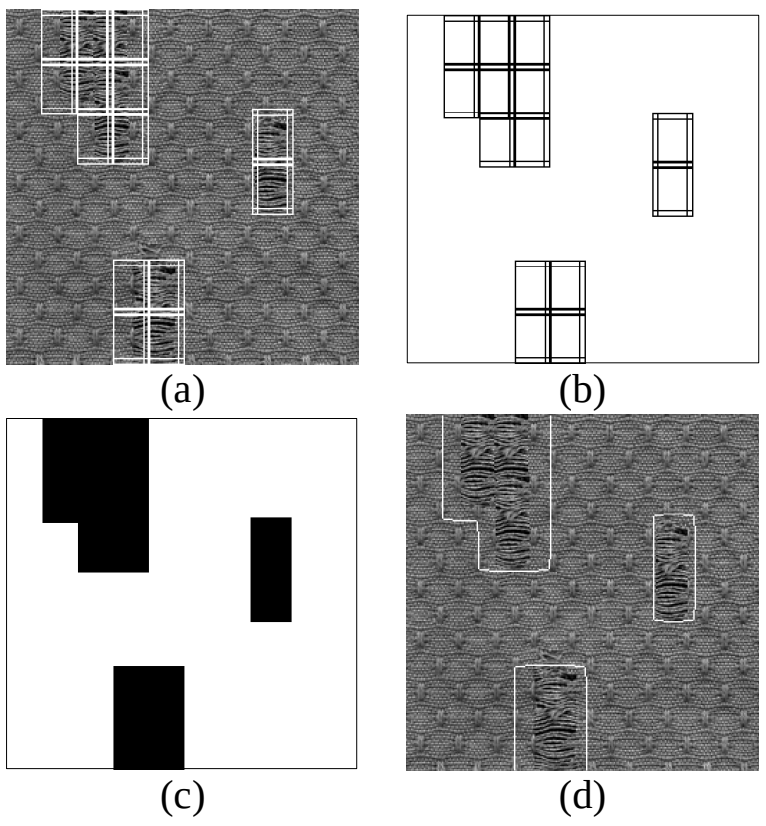

Figure 5. Illustration of defect fusion: (a) Boundaries of the defective blocks identified from each cropped image shown superimposed on the original image; (b) Boundaries of the defective blocks shown separately on plain background; (c) Result of morphological filling; (d) Canny edge identified shown superimposed on original defective image using white pixels.

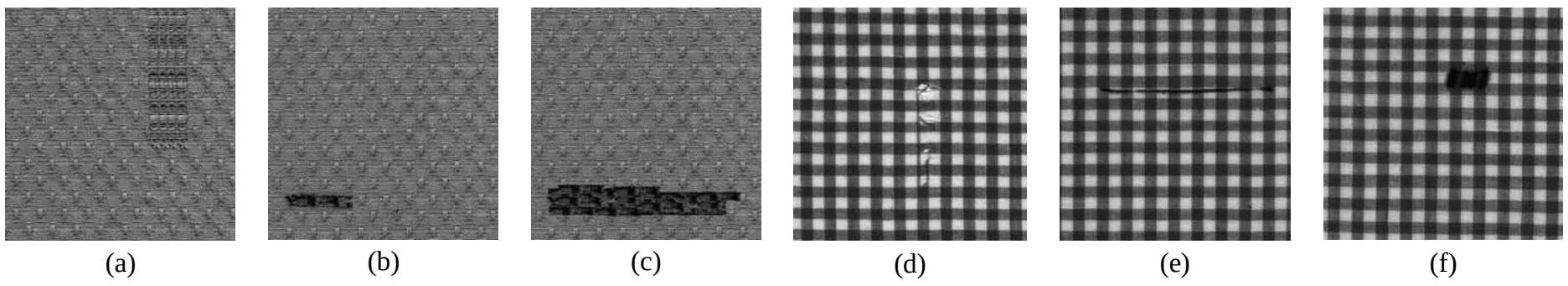

Figure 6. Sample real fabric images with defects: (a), (b), and (c) represent pmm images with defects - BE, TNB and TKB respectively; (d), (e) and (f) represent p4m images with defects - BE, TNB and TKB respectively.

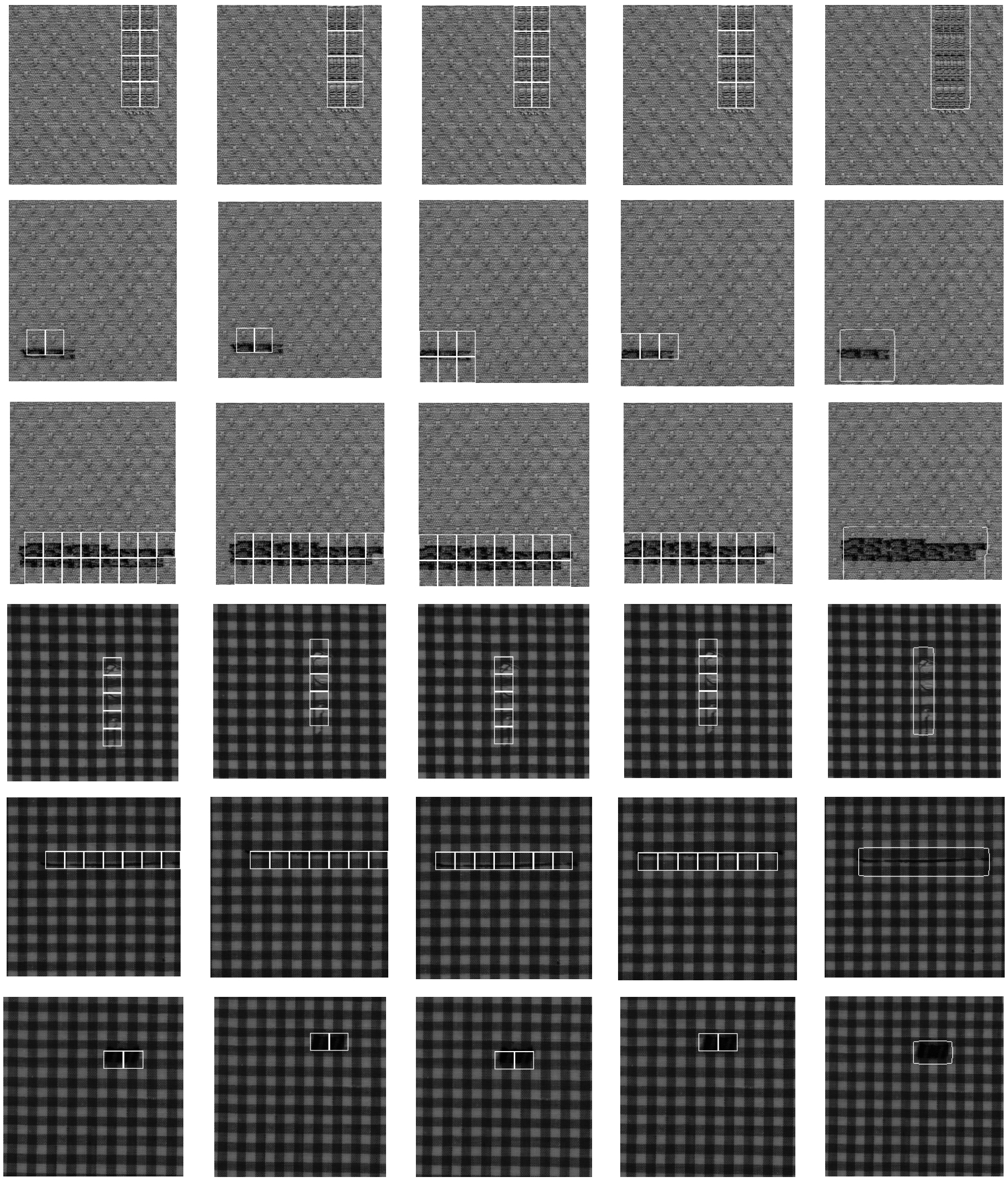

Figure 7. Defect detection on real fabric images: First, second, third and fourth columns show the identified defective periodic blocks from cropped images obtained from top-left, bottom-left, top-right, and bottom-right corners of the defective images; Fifth column shows the final result after merging of defects, morphological filling and edge detection; First, second, third, fourth, fifth and sixth rows show the defect detection result for pmm image with defect – BE, pmm image with defect – TNB, pmm image with defect – TKB, p4m image with defect – BE, p4m image with defect – TNB, and p4m image with defect – TKB respectively. It may be noted that since the box patterns in p4m images are too bright, gray values in all p4m images are linearly scaled down by a factor 0.5 so as to make the boundaries of the defects appear with better clarity to viewers.

## About the authors

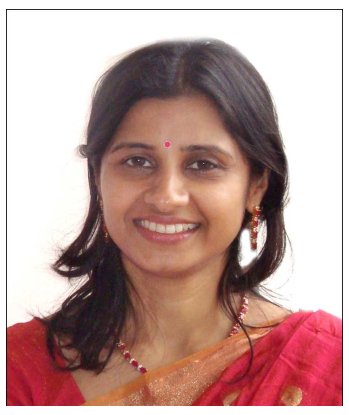

**V. Asha** received her Bachelor and Master Degrees in Computer Science from Mysore University during 1991 and 1993 respectively. Currently, she is working as Assistant Professor in Department of Computer Applications, New Horizon College of Engineering, Bangalore, Karnataka, India. Her research area includes image processing and pattern recognition.

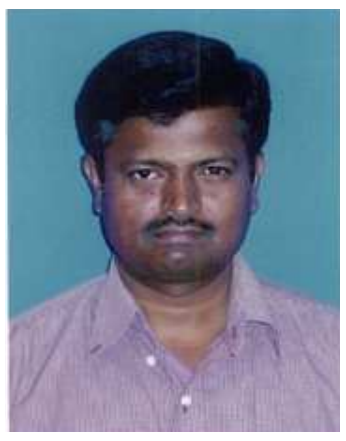

**Dr. N. U. Bhajantri** received his M.Tech and Ph.D from Indian Institute of Technology Delhi and Ph.D from Mysore University during 1999 and 2007 respectively. Currently, he is a professor in Govt. Engineering College, Chamarajanagar, Mysore District, Karnataka, India. He works in the broad area of image processing, pattern recognition and other related fields.

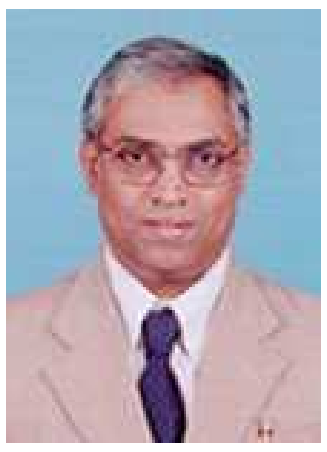

**Dr. P. Nagabhushan** received his M.Tech and Ph.D from Birla Institute of Technology, Mesra, Ranchi, and Mysore University during 1983 and 1988 respectively. Currently, he is in Department of Computers and Studies of Mysore University, Mysore, Karnataka, India. He works in the broad area of Cognition-Recognition, Pattern Recognition, Digital Image Processing, Object recognition, Data and Knowledge Mining, Document Image Analysis and other related fields.